\documentclass[A4paper, 10pt, conference]{ieeeconf}  
\IEEEoverridecommandlockouts       
\usepackage{geometry}
\usepackage{graphics} % for pdf, bitmapped graphics files
\usepackage{epsfig} % for postscript graphics files
\usepackage{mathptmx} % assumes new font selection scheme installed
\usepackage{times} % assumes new font selection scheme installed
\usepackage{amsmath} % assumes amsmath package installed
\usepackage{amssymb}  % assumes amsmath package installed
\usepackage{subfigure}
\usepackage{algorithm} %format of the algorithm
\usepackage{algorithmic} %format of the algorithm
\usepackage{multirow} %multirow for format of table
\usepackage{float}    % 如果使用 [H]，需要加载这个包

\usepackage{xcolor}

\begin{document}

% Change to your title
\title{\LARGE \bf
Depth-aware Fusion Method based on Image and 4D Radar Spectrum for 3D Object Detection
}

\author{Yue Sun$^{1}$, Yeqiang Qian$^{*, 2,3}$, Chunxiang Wang$^{2,3}$ and Ming Yang$^{2,3}$ % <-this % stops a space
\thanks{
	$\dag$ This work is supported by the National Natural Science Foundation of China (62103261/62173228/62373250).
}
\thanks{
	$^{1}$ The Global Institute of Future Technology, Shanghai Jiao Tong University, Shanghai, 200240, China.
%   Yeqiang Qian is the corresponding author and with Department
%of Engineering, XXX University, Austin, TX, USA, {\tt\small xxx@xxx.edu}. Eric Fuhlbrigge is with High Corporate Research Center, High Inc., CT, USA.%%
}
\thanks{
	$^{2}$ The Department of Automation, Shanghai Jiao Tong University, Shanghai, 200240, China.
}
\thanks{
	$^{3}$ Key Laboratory of System Control and Information Processing, Ministry of Education of China, Shanghai, 200240, China.
}
\thanks{
	$^{*}$ Corresponding author: Yeqiang Qian {\tt\small qianyeqiang@sjtu.edu.cn}.
}
}

\maketitle 
\thispagestyle{empty}

\begin{abstract}
Safety and reliability are crucial for the public acceptance of autonomous driving. 
To ensure accurate and reliable environmental perception, 
intelligent vehicles must exhibit accuracy and robustness in various environments. 
Millimeter-wave radar, known for its high penetration capability, 
can operate effectively in adverse weather conditions such as rain, snow, and fog. 
Traditional 3D millimeter-wave radars can only provide range, Doppler, and azimuth information for objects. 
Although the recent emergence of 4D millimeter-wave radars has added elevation resolution, 
the radar point clouds remain sparse due to Constant False Alarm Rate (CFAR) operations. 
In contrast, cameras offer rich semantic details but are sensitive to lighting and weather conditions. 
Hence, this paper leverages these two highly complementary and cost-effective sensors, 4D millimeter-wave radar and camera. 
By integrating 4D radar spectra with depth-aware camera images and employing attention mechanisms, 
we fuse texture-rich images with depth-rich radar data in the Bird’s Eye View (BEV) perspective, enhancing 3D object detection. 
Additionally, we propose using GAN-based networks to generate depth images from radar spectra in the absence of depth sensors, 
further improving detection accuracy.
\end{abstract}

% ================================================================================================
\section{Introduction}
Safety and reliability are key factors for the widespread acceptance of autonomous driving technology. 
Hence, developing robust all-weather environmental perception algorithms is essential. 
Common sensors used in autonomous driving for 3D object detection include cameras, LiDAR, and millimeter-wave radar. 
Cameras can provide rich visual information such as color and texture but are sensitive to lighting and weather. 
LiDAR offers dense point clouds but faces limitations in adverse weather and is high-cost. 
Millimeter-wave radar performs exceptionally well under challenging conditions, is cost-effective, 
and can measure speed and detect objects at long ranges, 
making it promising for autonomous driving \cite{4dsurvey}. 
However, its main challenge is the sparsity of measurements, 
exacerbated by traditional processing methods like Side Lobe Suppression (SLS)  \cite{sls} and Constant False Alarm Rate (CFAR) \cite{cfar}, 
leading to severe information loss. 
Recent advancements in 4D millimeter-wave radar provide elevation information but still suffer from low data density and noise.

To achieve accurate and reliable perception, 
autonomous vehicles typically fuse data from multiple sensor modalities for environmental perception and 3D object detection. 
Fusing signals from cameras and millimeter-wave radars enables achieving a comprehensive and
cost-effective perception of the surrounding environment in terms of contours, color, texture, distance, and speed. 
Additionally, the fusion system can operate continuously under all weather conditions and varying light levels. 
However, challenges remain in fusing camera and radar signals \cite{dpft}.  
Millimeter-wave radar data can be represented in various data formats depending on the processing stage, 
including raw ADC signals, 
radar spectra after applying Fast Fourier Transform (FFT) on ADC signals, % along each dimension (range, doppler, azimuth, and elevation), 
and radar point clouds after further applying SLS and CFAR operations \cite{exploringradardatarepresentations}. 
Currently, radar point clouds are mostly used in 3D object detection algorithms 
but face issues due to their high sparsity and inherent differences from LiDAR point clouds, 
making them less effective for detection tasks. 
Moreover, there are challenges related to camera-radar sensor fusion in terms of resolution variations and perceived dimensionality \cite{dpft}.

\begin{figure*}[ht!]
	\centerline{
		\includegraphics[width=1\textwidth]{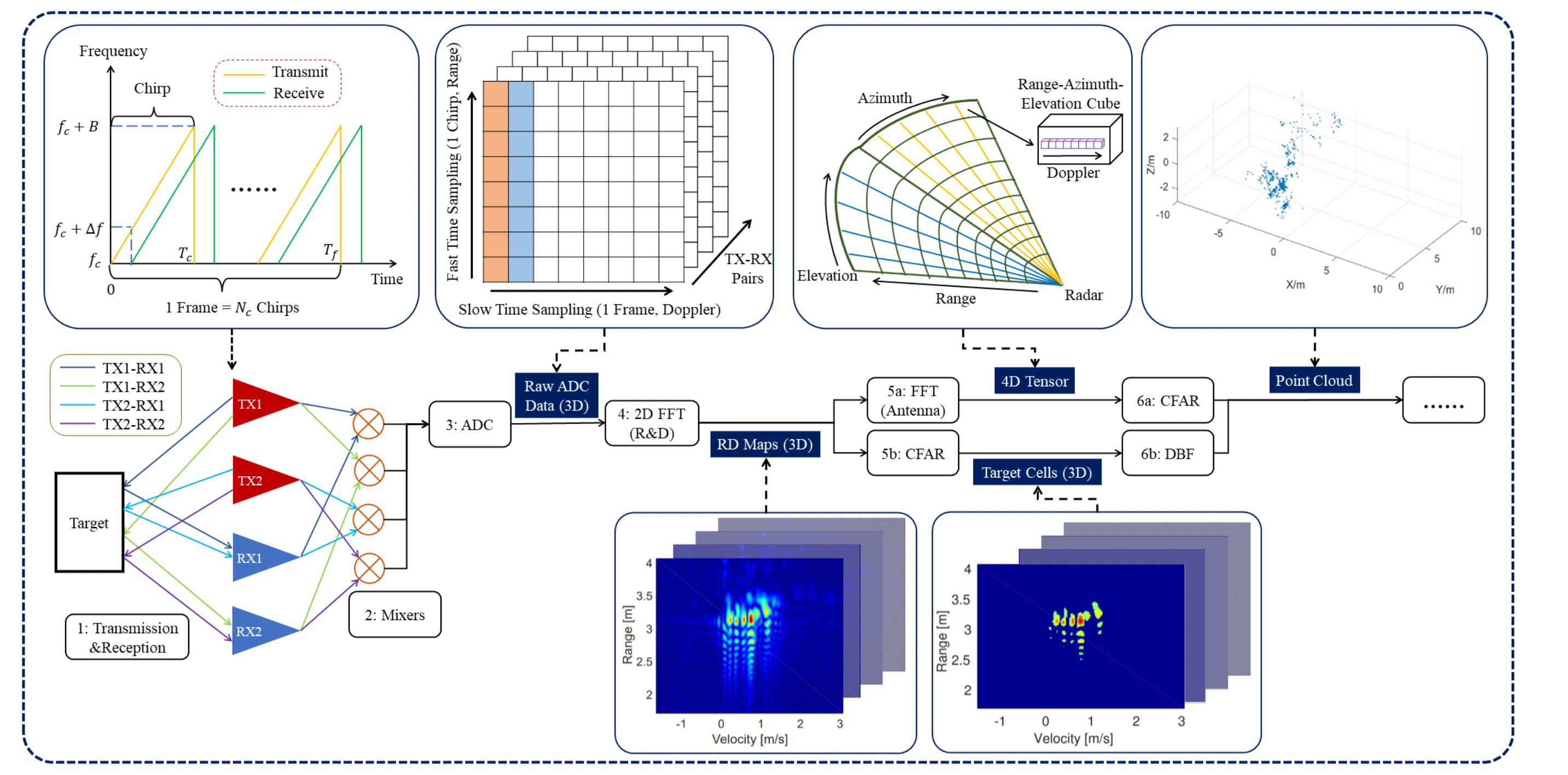}
	}
	\caption{Traditional signal processing pipeline of 4D millimeter-wave radar data and the corresponding data format after each step \cite{4dsurvey}.
	}
	\label{radar-pipeline}
\end{figure*}

To address the aforementioned issues, 
this paper proposes using raw 4D radar spectra and depth-aware images as input data. 
Features are extracted separately from the RGB and depth images to obtain corresponding multi-scale features, 
which are then fused to produce multi-scale image features with depth information. 
Following a similar approach to EchoFusion \cite{echofusion}, 
image and radar features are fused in the BEV feature space under polar coordinates 
based on column-wise attention for image features and row-wise cross-attention for radar features, 
aggregating high resolution of image features in the elevation direction and radar features in the depth direction. 
The resulting BEV features and object queries are then applied in the polar decoder \cite{polarformer} and detection head for 3D object predictions.

The main contributions of this paper are as follows:
\begin{enumerate}
	\item 
	 We propose an all-weather object detection algorithm that integrates depth-aware images and 4D millimeter-wave radar spectra in the BEV feature space 
	 to achieve optimal 3D object detection performance.
	\item 
	We introduce a method to augment features from RGB images with depth images before camera-radar data fusion, 
	and we propose using a GAN-based depth generator to produce depth images from radar spectra when a depth sensor is unavailable.
	\item 
	Our model achieves good detection performance while significantly reducing network complexity.
\end{enumerate}

% ================================================================================================
\section{Related Work}

\subsection{Radar Data Representations}
Millimeter-wave radar uses radio waves in the millimeter-wave frequency range (30 GHz to 300 GHz) for environmental sensing. 
Typically, each transmitting antenna (TX) emits a sequence of Frequency Modulated Continuous Waves (FMCW), called chirps, 
with linearly increasing frequency \cite{texas}. 
Reflected waves from objects are captured by receiving antennas (RX) and processed to determine distance, speed, and angle. 
Traditional 3D radars, with horizontally arranged antennas, 
provide only horizontal detection and lack elevation information. 
In contrast, 4D radars use both horizontal and vertical antenna arrays, 
enabling the measurement of elevation angles as well. 
4D radar data can be represented as ADC signals, radar spectra, or point clouds, 
depending on the processing stage. 
%This section will review the typical 4D millimeter-wave radar signal processing workflow to illustrate how these data representations are generated.

Fig. \ref{radar-pipeline} shows the traditional process of converting 4D millimeter-wave radar signals 
from raw reflected waves to radar point clouds, 
along with the corresponding data format at each step.
\begin{itemize}
	\item Step 1: Each radar TX sends chirps, and each RX receives the reflected signals.
	\item Step 2: The received signal and its corresponding transmitted signal are mixed 
							and passed through a low-pass filter to generate an Intermediate Frequency (IF) signal.
	\item Step 3: The IF signal is converted into raw ADC data by an Analog-to-Digital Converter.
	\item Step 4: Apply FFT to the ADC data along the range and Doppler dimensions to produce a series of Range-Doppler (RD) maps 
						corresponding to different TX-RX pairs.
	\item Step 5 \& 6: One approach is to apply FFT along the azimuth and elevation dimensions of different TX-RX pairs
									to obtain a 4D radar spectrum (Step 5a). 
									Then, in Step 6a, CFAR operation is applied to the four-dimensional data to filter it based on intensity, 
									resulting in a radar point cloud. 
									Another approach is to first apply a CFAR-type algorithm to filter the RD map and generate target cells (Step 5b), 
									and then use Digital Beamforming (DBF) in Step 6b to recover angle information and generate the radar point cloud.
\end{itemize}

\subsection{Camera-Radar Datasets}
Although many public datasets now include millimeter-wave radar data alongside images and LiDAR data, 
these radar datasets contain fewer samples compared to the latter two. 
Most existing datasets, such as NuScenes \cite{nuscenes}, RADDet \cite{RADDet}, Zendar \cite{zendar}, RADIATE \cite{RADIATE}, and CARRADA \cite{CARRADADC}, 
use traditional 3D radar, 
providing data along the range, Doppler, and azimuth dimensions but lacking elevation information. 
This limitation makes accurate 3D bounding box estimation challenging in 3D object detection.
With the emergence of 4D millimeter-wave radar, 
more public datasets have begun incorporating 4D radar to capture data with elevation information. 

However, most of these datasets provide radar point clouds processed by methods like CFAR, 
such as View-of-Delft \cite{vod}, Astyx \cite{astyx}, and RadarScenes \cite{radar_scenes}. 
Due to the sparsity of radar signals, we aim to use 4D radar that provides elevation information, 
and the radar data should be in the form of lossless \textit{raw radar data} before CFAR operations, 
including ADC data and radar spectra. 
Among available datasets, RADIal and K-Radar meet these criteria. 
However, RADIal initially only offered 2D annotations, 
and although Liu et al. \cite{echofusion} recently added 3D annotations, 
it remains limited in scope and lacks data under adverse weather conditions. 
Therefore, the K-Radar dataset is the most suitable for this research.

K-Radar contains 35K frames of 4D radar spectrum/tensor (4DRT) with measurements across 
the range, Doppler, azimuth, and elevation dimensions, 
alongside multi-view images, high-resolution LiDAR point clouds, annotated 3D bounding boxes, and other relevant data. 
K-Radar also covers various challenging road structures (e.g., urban, suburban roads, alleyways, and highways) 
and adverse weather conditions (e.g., fog, rain, and snow) \cite{k-radar}.

\begin{figure*}[ht]
	\centerline{
		\includegraphics[width=1\textwidth]{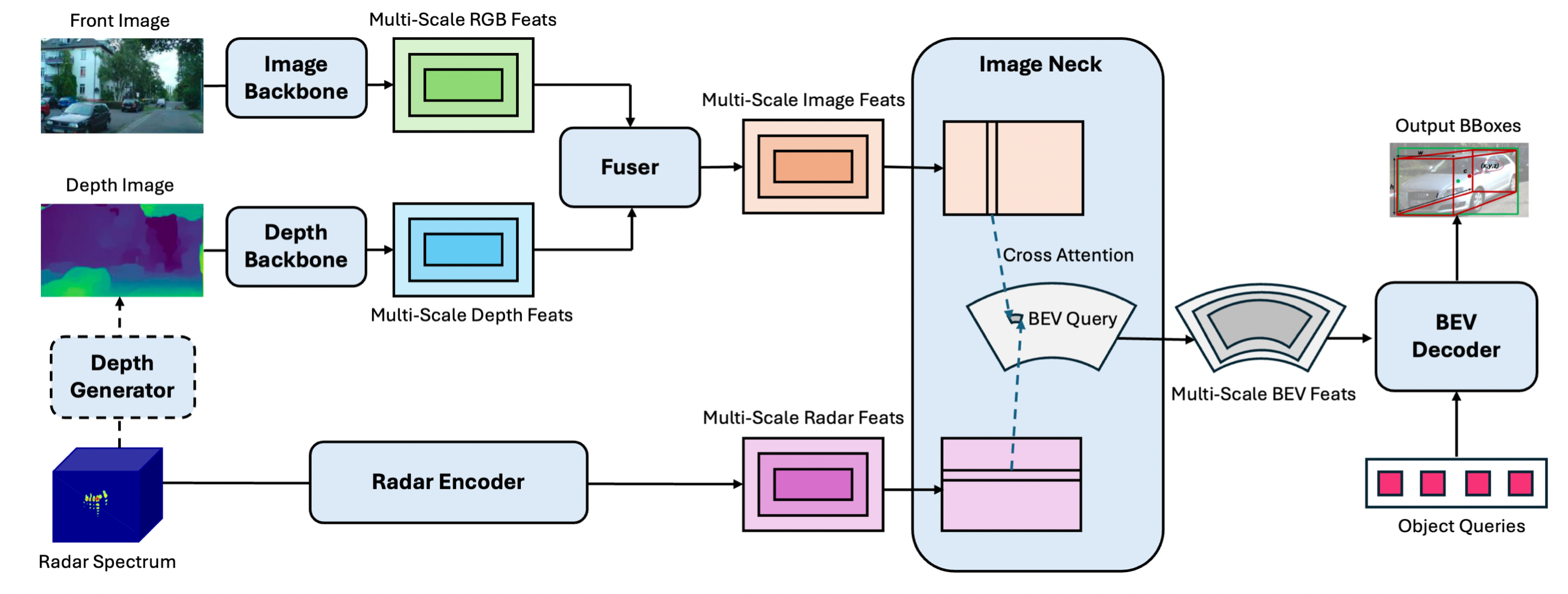}
	}
	\caption{The overall architecture of our proposed algorithm.}
	\label{our_arch}
\end{figure*}

\subsection{Camera-Radar Fusion based 3D Object Detection}
Previous radar-camera fusion-based 3D object detection algorithms often rely on traditional 3D millimeter-wave radar signals, 
which are typically represented as radar point clouds. 
Methods like CenterFusion \cite{centerfusion}, RCBEVDet \cite{rcbevdet}, and BEVFusion \cite{bevfusion} 
extract radar features similar to LiDAR and then fuse them with image features through concatenation or cross-attention before feeding them into the detection network. 
However, these traditional 3D radar point clouds lack elevation information and lose significant data due to CFAR thresholding, 
which negatively impacts the model’s detection performance. 
As a result, researchers often prefer high-resolution sensors like cameras and LiDARs for multisensor fusion to achieve better detection performance.

With the emergence of 4D millimeter-wave radars, 
many studies have begun fusing more cost-effective and highly complementary cameras and 4D millimeter-wave radars in 3D object detection algorithms.
Methods such as LXL \cite{lxl} and RCFusion \cite{rcfusion} 
fuse 4D radar point clouds with image features by first pillarizing the radar point clouds to extract BEV features, 
which are then fused with image features in the BEV feature space.
To mitigate the information loss inherent in radar point clouds, 
EchoFusion \cite{echofusion} and DPFT \cite{dpft} propose fusing raw radar data with camera images. 
However, EchoFusion has high computational resource demands (requiring 4.0 GiB of GPU memory during the inference stage \cite{dpft}), 
while DPFT uses a full 4D tensor as input, which requires significant memory, and is difficult to obtain \cite{k-radar}. 
Building on EchoFusion's work, 
our method enhances image features with depth information before radar-camera fusion, 
which can improve 3D object detection performance while reducing network complexity. 
Additionally, we propose that depth information can still be derived from 4D radar spectra even in the absence of a depth sensor.

% ================================================================================================

\section{Methodology}\label{method}

\subsection{System Overview}
Our overall architecture is illustrated in Fig. \ref{our_arch}. 
It takes depth-aware images from a front camera and radar spectra from a 4D millimeter-wave radar as inputs. 
For the radar spectrum input branch, 
we use a convolutional neural network (CNN) to extract multi-scale radar features. 
For the image input branch, 
we first separately extract multi-scale features from RGB images and depth images, 
and then fuse these features to obtain multi-scale image features with depth information. 
In the BEV space under polar coordinates, 
polar queries at each scale fuse image features and radar features at that scale through an attention mechanism to produce multi-scale BEV features. 
Finally, the multi-scale BEV features and object queries are fed into the polar decoder \cite{polarformer} 
and detection heads to estimate the classes and bounding boxes of 3D objects.

\subsection{RGB and Depth Images Fusion}
Since each pixel value in a depth image represents the spatial depth information of the corresponding location of the object, 
while each pixel value in an RGB image corresponds to the color and texture information, 
feature extraction should be performed separately for each. 
We preprocess the single-channel depth image by expanding it to three channels through simple repetition.
Then, we use the lightweight convolutional network ResNet18 \cite{resnet} as the backbone 
to extract multi-scale features from both RGB and depth images. 
The resulting multi-scale features are concatenated along the channel dimension and 
fed into a Feature Pyramid Network (FPN) \cite{fpn} for feature fusion, 
which outputs multi-scale image features with depth information.

\subsection{Image and Radar Feature Fusion}
For the BEV feature space at the $l-level$ in the polar coordinate system, 
we first initialize it uniformly with $R^l \times A^l$ BEV queries,
where $R^l$ and $A^l$ are the range and azimuth resolutions of the BEV features at level $l$,
and each query is represented as $q^l (r_{bev}, \phi_{bev})$.

The fusion of image feature $F_I^l$ and radar feature $F_R^l$ is based on the fact that 
each polar column with the same azimuth $ \phi_{bev}$ in the BEV features 
corresponds to a column $x_I^l$ in the image feature, 
and each polar row with the same range $ r_{bev}$ in the BEV features 
corresponds to a row in the radar feature $R_r^l$. 
Following the polar-aligned Attention (PAA) technique proposed in EchoFusion \cite{echofusion},
we fuse the image and radar features as follows:

We use cross-attention to update the BEV queries with image features as follows:
\begin{equation}
	 \hat{Q}^l = CrossAttention(Q_1^l, K_1^l, V_1^l) 
\end{equation}
%\begin{equation}
%	\left\{
%	\begin{aligned}
%		& Q_1^l = \text{all }q^l (r_{bev}, \phi_{bev})\text{ with the same }\phi_{bev}\\
%		& K_1^l = \text{all }f^l_I(x_I, y_I)\text{ with the same }x_I\\
%		& V_1^l = \text{all }f^l_I(x_I, y_I)\text{ with the same }x_I
%	\end{aligned}
%	\right.
%\end{equation}
% --------------------------------
\begin{equation}
	 Q_1^l = \text{all }q^l (r_{bev}, \phi_{bev})\text{ with the same }\phi_{bev}
\end{equation}
\begin{equation}
K_1^l = V_1^l = \text{all }f^l_I(x_I, y_I)\text{ with the same }x_I\\
\end{equation}
% --------------------------------

Then, we update the BEV queries with radar features based on:
\begin{equation}
	\tilde{Q}^l = CrossAttention(Q_2^l, K_2^l, V_2^l) 
\end{equation}
\begin{equation}
Q_2^l = \text{all }\hat{q}^l (r_{bev}, \phi_{bev})\text{ with the same }r_{bev}
\end{equation}
\begin{equation}
K_2^l = V_2^l = \text{all }f^l_R(a_I, r_I)\text{ with the same }r_I
\end{equation}

\subsection{Depth Image Generation From Radar Spectrum} \label{depth_generation}

\begin{figure}[ht!]
	\centering
	\centerline{
		\includegraphics[width=0.49\textwidth]{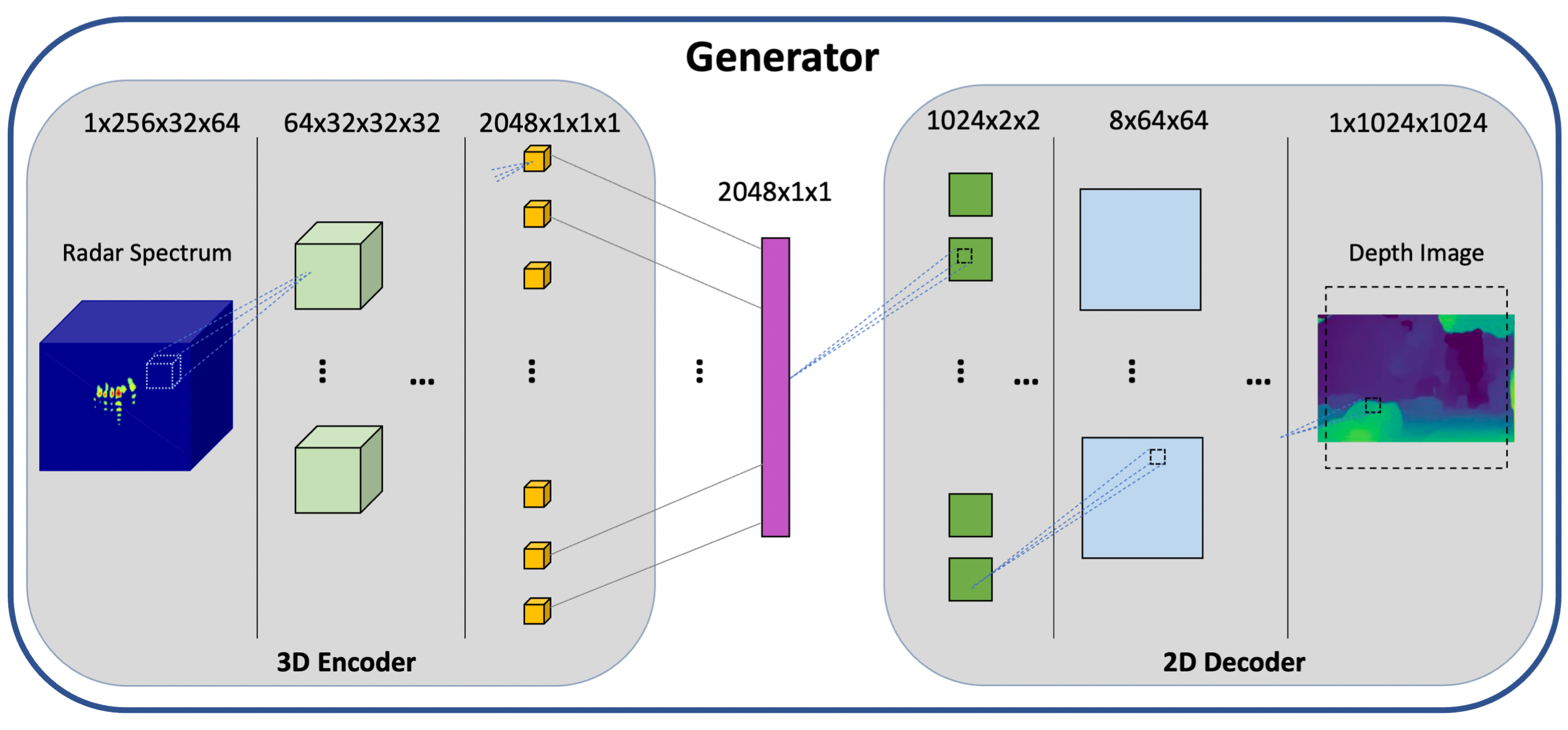}
	}
	\caption{The network architecture of our depth image generator based on the design proposed in \cite{3drimr}.
	}
	\label{depth_generator}
\end{figure}

Depth images are vital for improving the accuracy of vision-based object detection \cite{rgbd_survey}. 
While depth sensors usually provide these images, they are sometimes absent in autonomous vehicles due to size and cost constraints. 
Inspired by work in \cite{3drimr, r2p, multi-obj}, 
we propose using a GAN-based depth generator to generate accurate depth images with high resolution from millimeter-wave radar spectra. 
By generating depth images before applying attention-based feature fusion, 
we can augment images with depth information, 
allowing image features to include depth information and thereby improve our model's overall detection performance. 
The depth generator network is similar to the one in \cite{3drimr}, 
except that skip connections are not used in the generator (as shown in Fig. \ref{depth_generator}).
The training data comes from K-Radar sequences 12-20, with ground truth depth obtained by projecting LiDAR point clouds onto images.

\subsection{Detection Head and Loss}
We use a similar architecture as used in \cite{echofusion} and \cite{polarformer}, 
but with the number of decoder layers reduced from 6 to 3, 
and we limit the object queries to 20 instead of 30 to minimize computation consumption. 
The resulting object queries are then fed into classification and regression detection heads 
to obtain the predicted categories \(c\) and 3D bounding boxes $\vec{b} = (x, y, z, l, w, h, \phi)$ of detected 3D objects. 

During model training, 
a Hungarian matching-based set-to-set loss is employed, 
which consists of Focal Loss for the classification network and L1 Loss for the regression network.

% ================================================================================================
\begin{figure*}[ht!]
%	\small
	\centering
	\begin{tabular}{c c c c c }
		% row 1 -------------------------------------------------------------------------------
		\begin{minipage}[b]{0.37\columnwidth}
			\centering
			\includegraphics[width=1.13\linewidth]{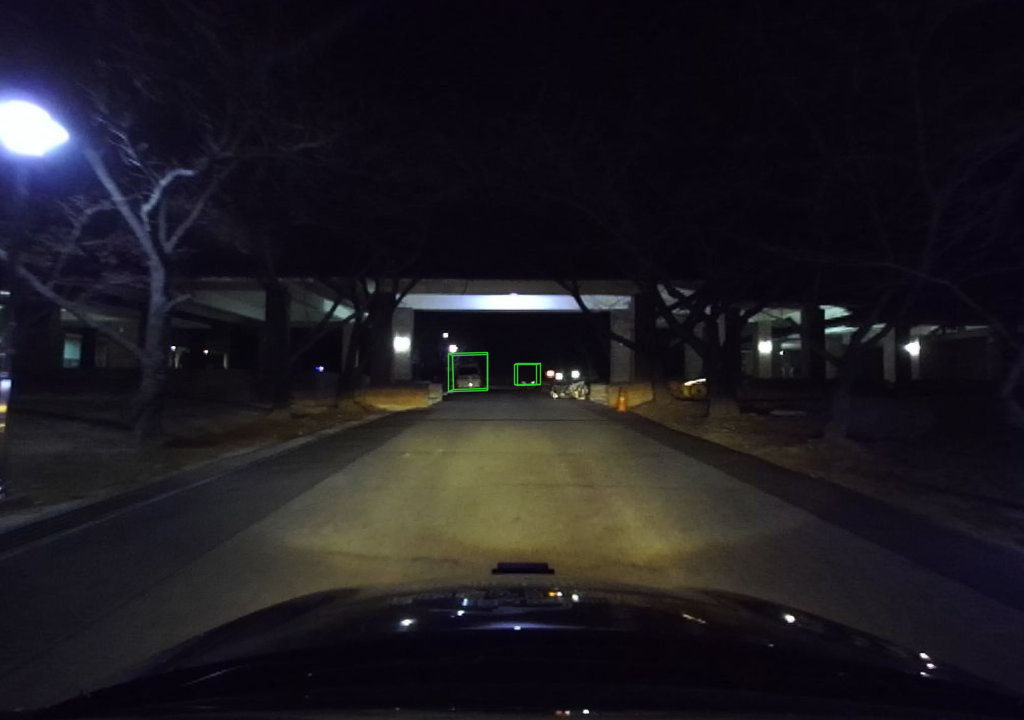}
		\end{minipage}
		&
		\begin{minipage}[b]{0.37\columnwidth}
			\centering
			\includegraphics[width=1.13\linewidth]{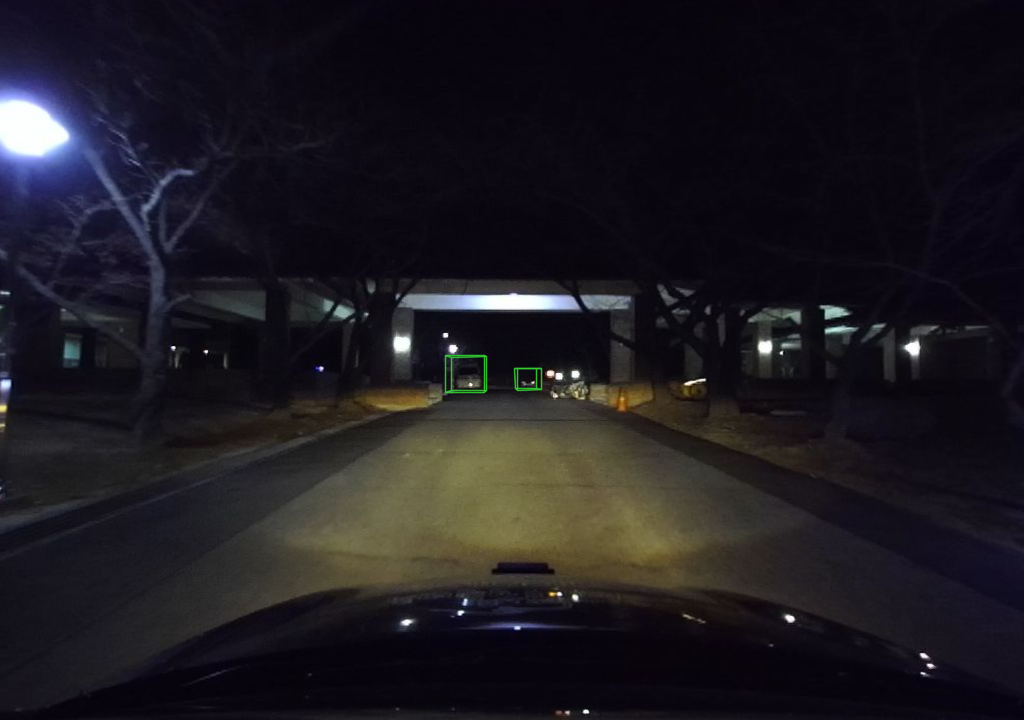}
		\end{minipage}
		&
		\begin{minipage}[b]{0.37\columnwidth}
			\centering
			\includegraphics[width=1.13\linewidth]{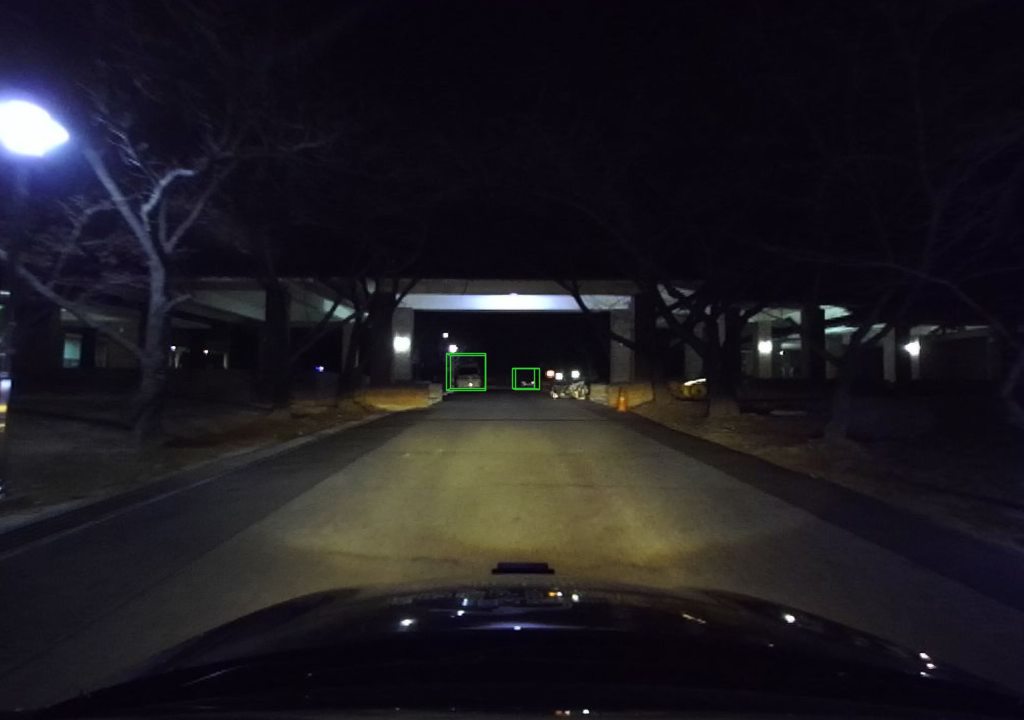}
		\end{minipage}
		&
		\begin{minipage}[t]{0.37\columnwidth}
			\centering
			\includegraphics[width=1.13\linewidth]{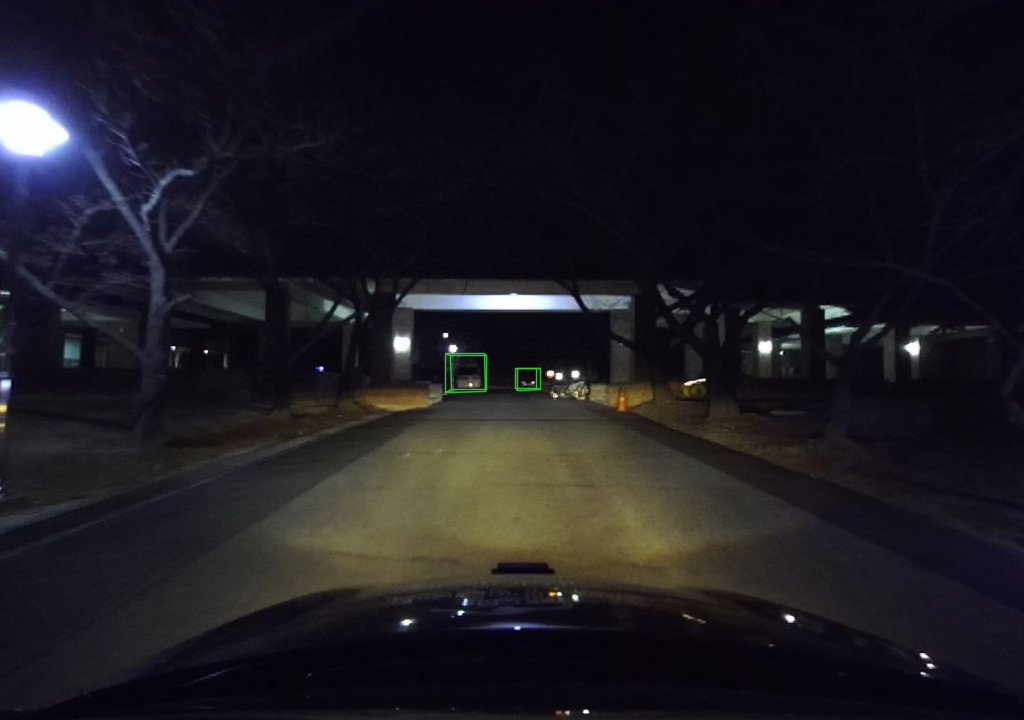}
		\end{minipage}
		&
		\begin{minipage}[t]{0.37\columnwidth}
			\centering
			\includegraphics[width=1.13\linewidth]{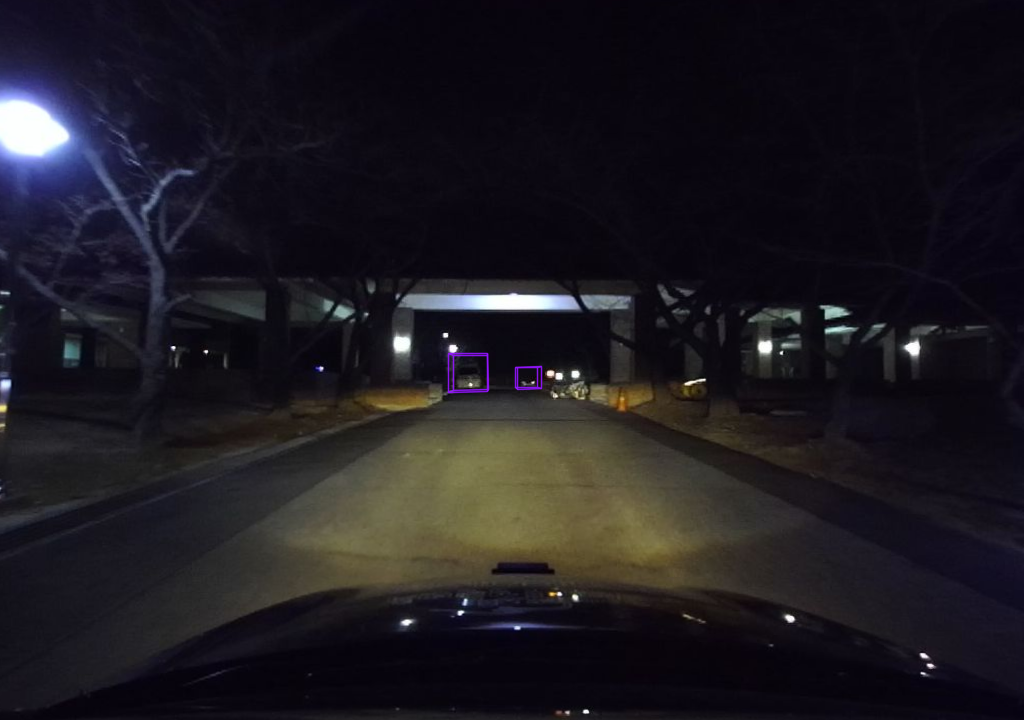}
		\end{minipage}
		\\
		% row 2 -------------------------------------------------------------------------------
		\begin{minipage}[b]{0.37\columnwidth}
			\centering
			\includegraphics[width=1.13\linewidth]{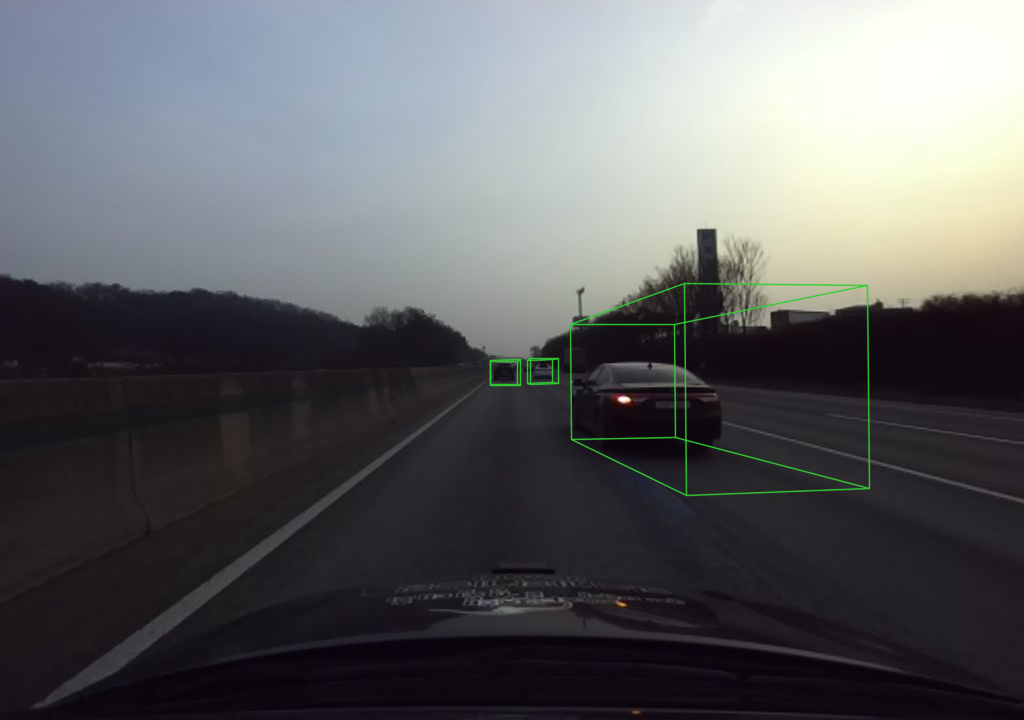}
		\end{minipage}
		&
		\begin{minipage}[b]{0.37\columnwidth}
			\centering
			\includegraphics[width=1.13\linewidth]{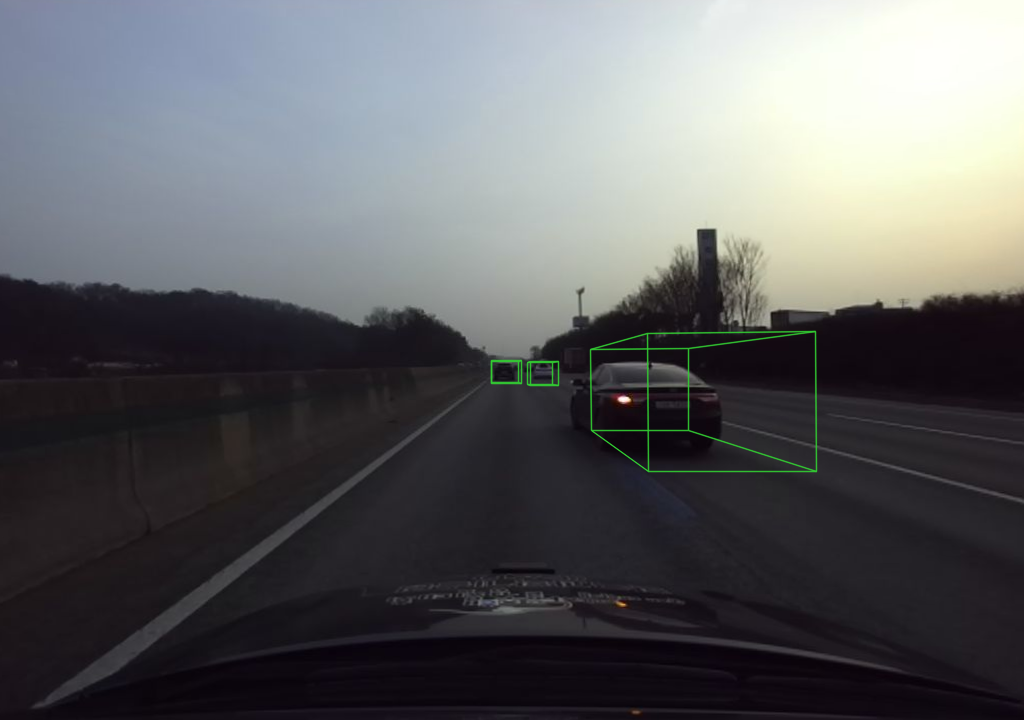}
		\end{minipage}
		&
		\begin{minipage}[b]{0.37\columnwidth}
			\centering
			\includegraphics[width=1.13\linewidth]{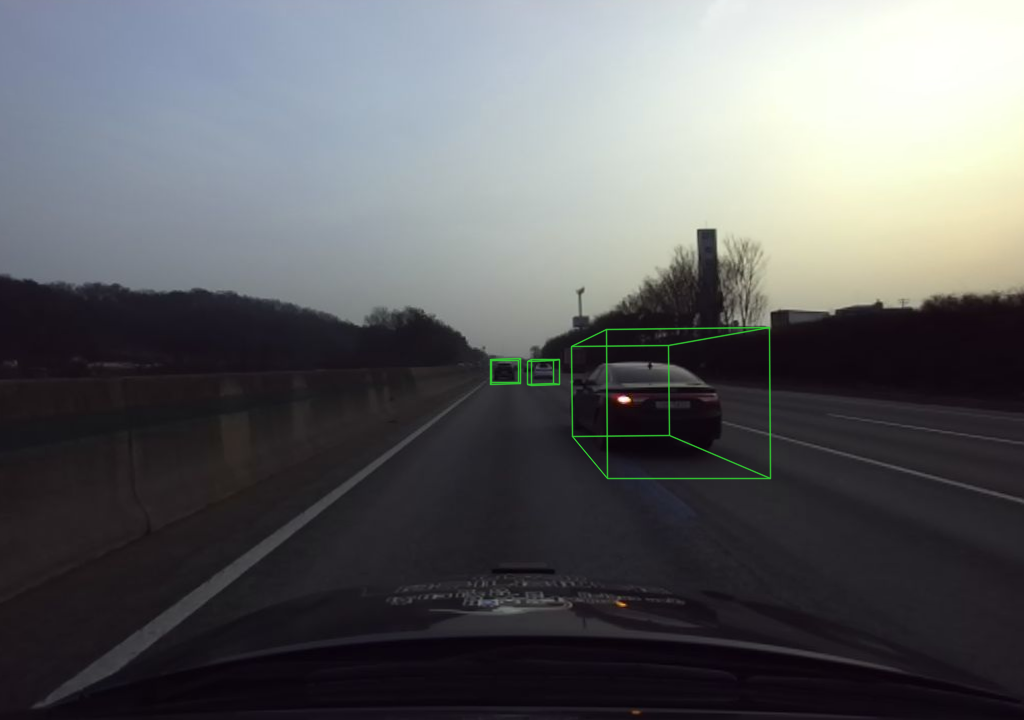}
		\end{minipage}
		&
		\begin{minipage}[t]{0.37\columnwidth}
			\centering
			\includegraphics[width=1.13\linewidth]{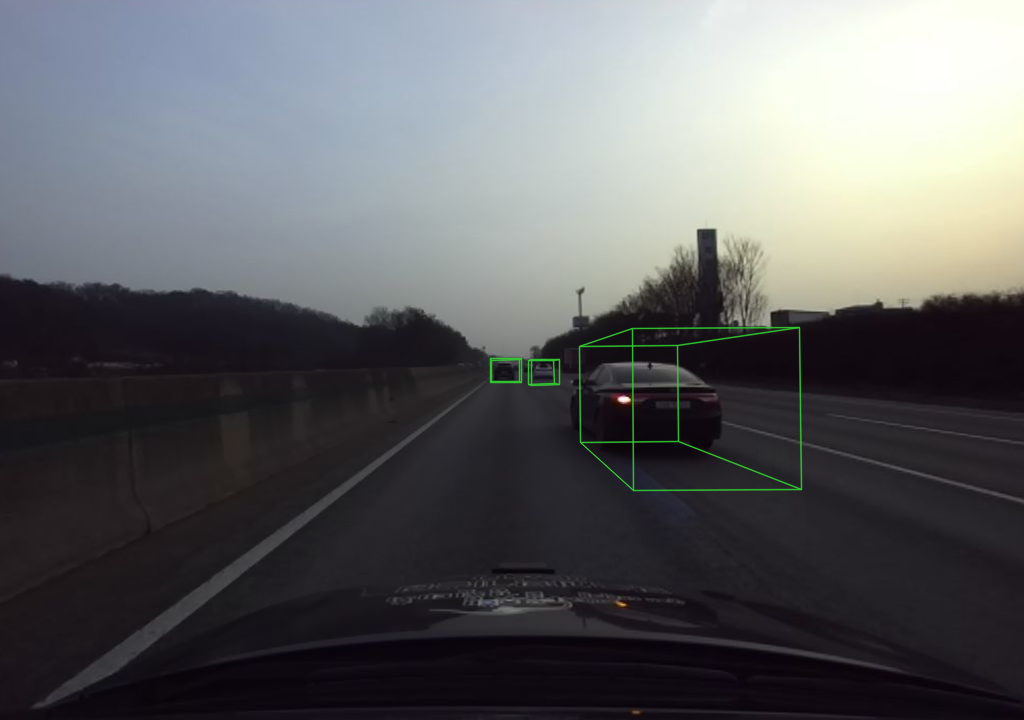}
		\end{minipage}
		&
		\begin{minipage}[t]{0.37\columnwidth}
			\centering
			\includegraphics[width=1.13\linewidth]{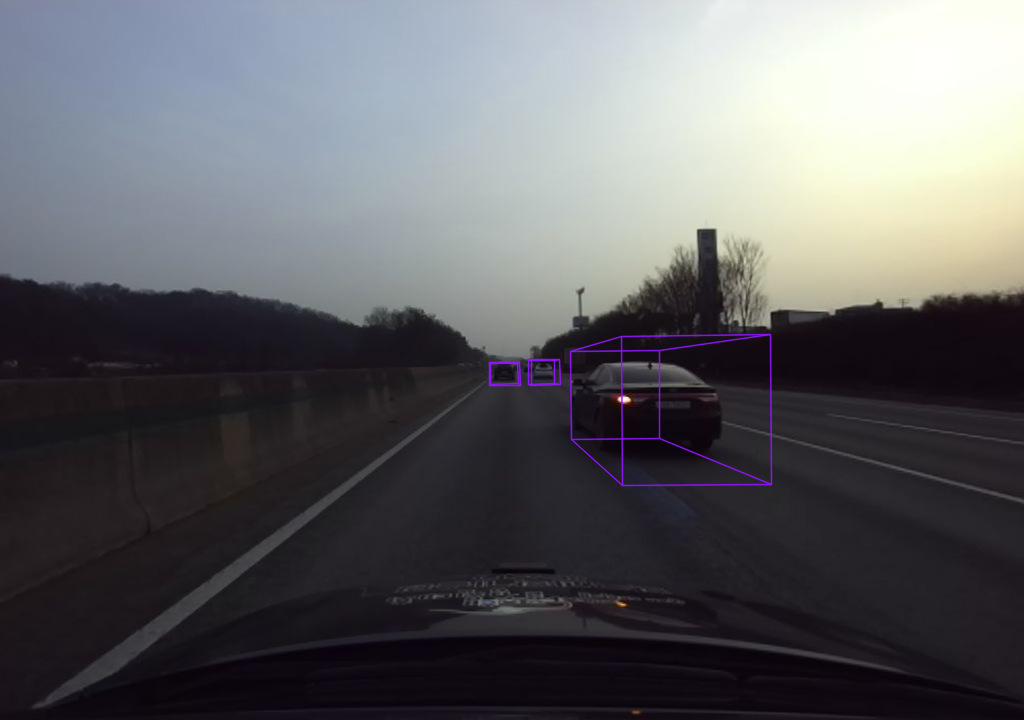}
		\end{minipage}
		\\
		% row 2 -------------------------------------------------------------------------------
		\begin{minipage}[b]{0.37\columnwidth}
			\centering
			\includegraphics[width=1.13\linewidth]{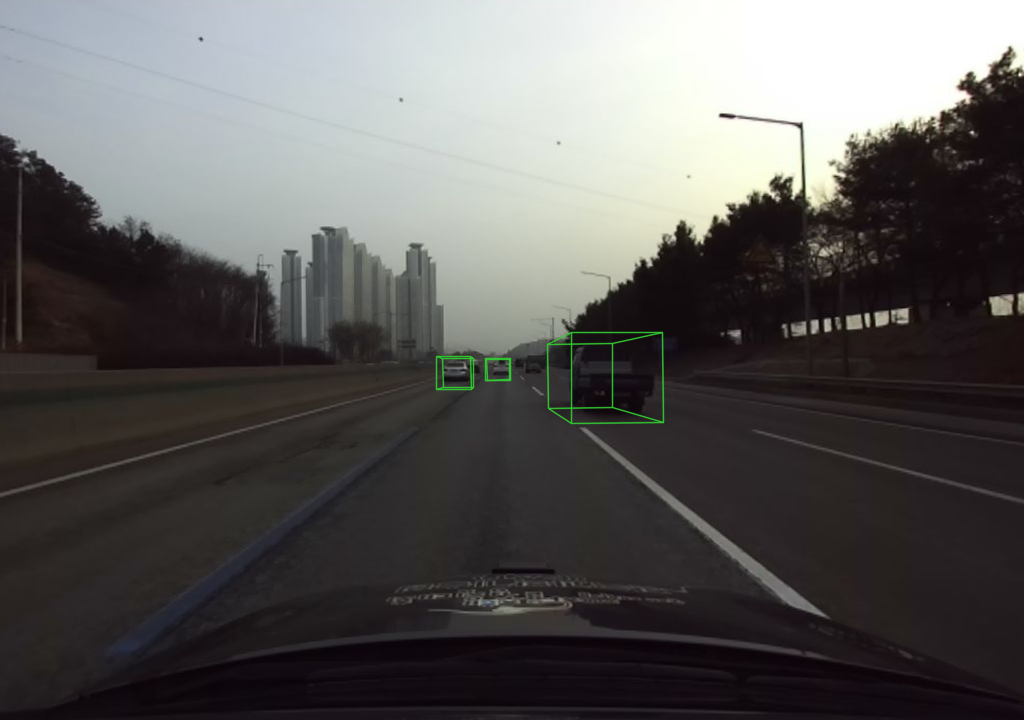}
		\end{minipage}
		&
		\begin{minipage}[b]{0.37\columnwidth}
			\centering
			\includegraphics[width=1.13\linewidth]{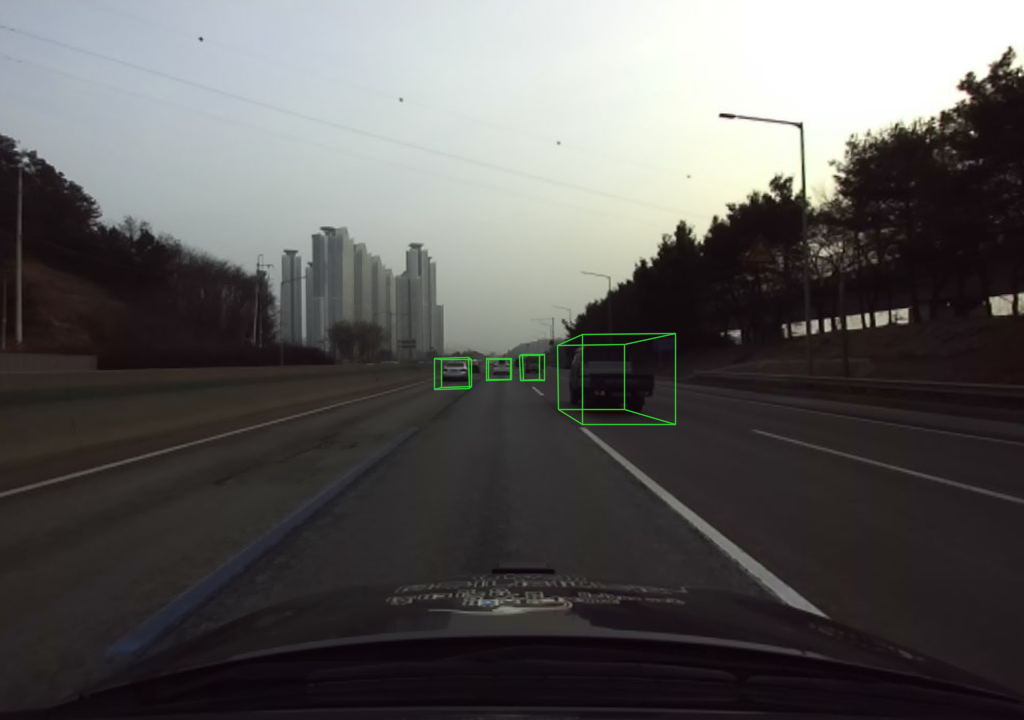}
		\end{minipage}
		&
		\begin{minipage}[b]{0.37\columnwidth}
			\centering
			\includegraphics[width=1.13\linewidth]{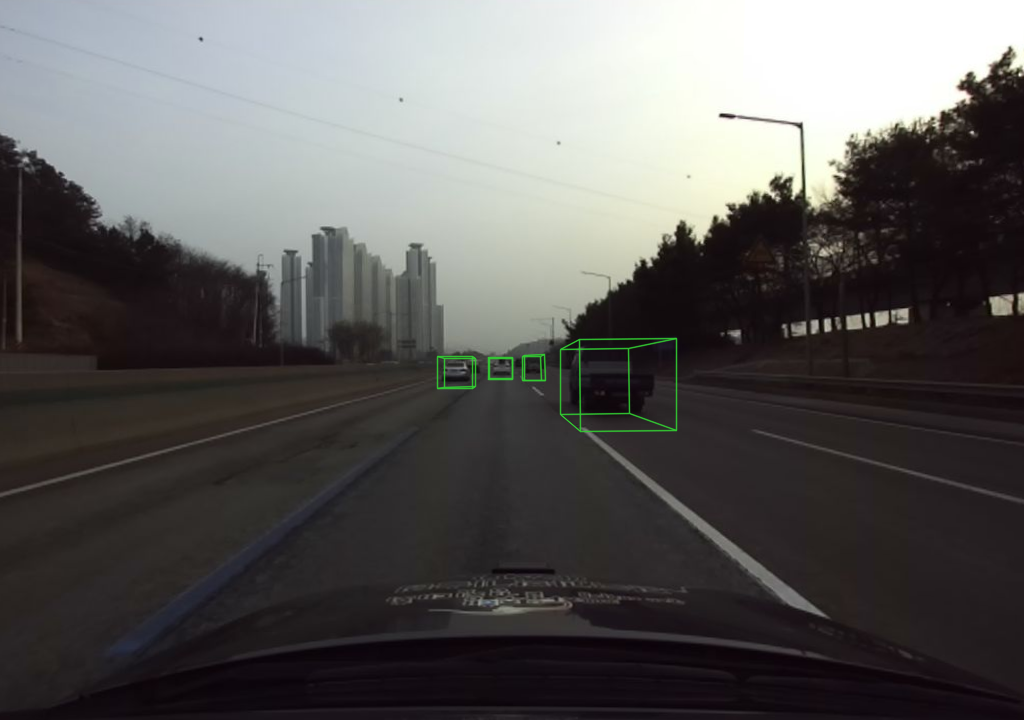}
		\end{minipage}
		&
		\begin{minipage}[t]{0.37\columnwidth}
			\centering
			\includegraphics[width=1.13\linewidth]{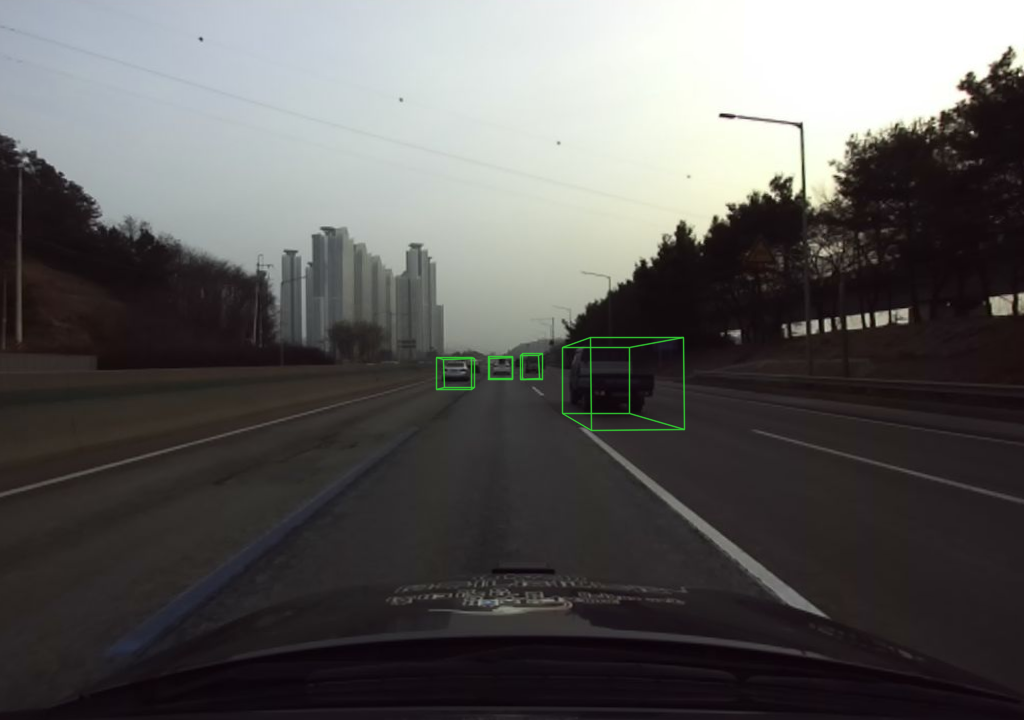}
		\end{minipage}
		&
		\begin{minipage}[t]{0.37\columnwidth}
			\centering
			\includegraphics[width=1.13\linewidth]{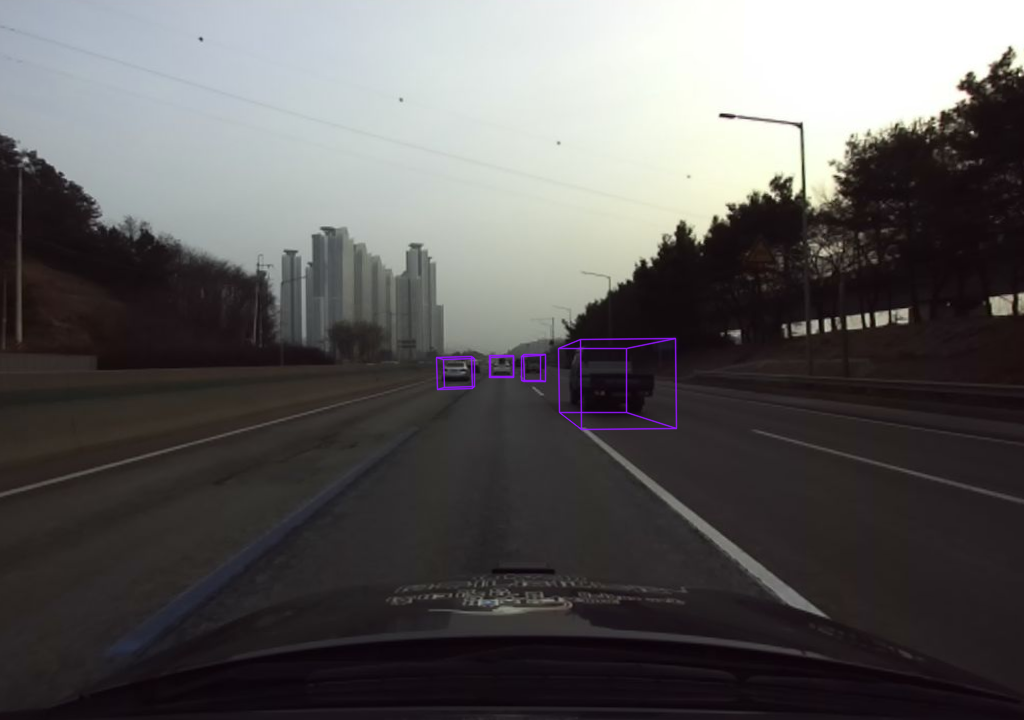}
		\end{minipage}
		\\
		% row: name ---------------------------------------------------------------------------------------
		EchoFusion
		&
		EchoFusion-small
		&
		Ours-depth64
		&
		Ours-depth128
		&
		GT
	\end{tabular}
	\caption{Visualization of outputs from different methods compared to ground truth.
%		EchoFusion-small: same architecture as Echofusion but with ResNe
	}
	\label{vis}
\end{figure*}

%00_s9_cam-front_01233.png

\section{Experiments}

\subsection{Dataset}
We use a subset of the K-Radar dataset \cite{k-radar}, 
specifically sequences 1-11, for our experiments.
This choice is due to the fact that only sequences 1-20 have been uploaded to Google Drive and are readily accessible.
Additionally, we initially reserved sequences 12-20 for training the depth generator.

\subsection{Model Training and Testing}\label{train_test}
We conduct all experiments on two NVIDIA GeForce RTX 3090 GPUs with a batch size of 4. 
Due to the limited size of our dataset, 
we reorganize the training and testing sets. 
A total of 7,755 frames from K-Radar sequences 1-11 are split into an approximately 9:1 ratio,
with 6,972 frames used for training and 783 frames for testing. 
We follow the same training strategy as used in \cite{echofusion}, 
but we train the model for only 10 epochs as we use about half the data they used for training.
Given the limited hardware resources and the relatively large network structure, 
we restrict our training and testing to the \textit{Sedan} category, 
which predominates in K-Radar sequences 1-11, 
accounting for over 77.3\% of all objects.

\subsection{Evaluation Results}
We evaluate the following models on our test dataset as mentioned in Subsection \ref{train_test}. 
\begin{itemize}
	\item \textit{EchoFusion \cite{echofusion}}: Configured with its default settings.
	\item \textit{EchoFusion-small}: A \textit{light} version of EchoFusion, 
														   featuring a \textit{smaller} network as used in our model
														   that replaces ResNet50 with ResNet18, 
														   reduces the number of decoder layers from 6 to 3, 
														   and utilizes only 20 object queries instead of 30. 
	\item \textit{Ours-depth64}: Our model, as described in Section \ref{method}, 
													 with the input depth images generated by projecting 64-line LiDAR point clouds onto the images.
	\item \textit{Ours-depth128}: The same as \textit{Ours-depth64}, but using 128-line LiDAR instead.
\end{itemize}

\begin{table}[h]
	\centering
	\renewcommand{\arraystretch}{1.5}
	\setlength{\tabcolsep}{4pt}
	\caption{Evaluation of detection performance using different methods. 
		%		IoU thresholds of 0.3, 0.5, and 0.7 are used. 
		%		The best result for each metric is highlighted in bold.
	}
	\begin{tabular}{|c||ccc|ccc|}
		\hline
		\multirow{2}{*}{Method} & \multicolumn{3}{c|}{ BEV AP (\%)} & \multicolumn{3}{c|}{3D AP(\%)} \\
												& 0.3                        & 0.5                        & 0.7                     & 0.3                      & 0.5                      & 0.7     \\ \hline \hline
		EchoFusion                   & 90.82                     & \textbf{90.73}     & \textbf{89.78}   & 90.79                  & 88.26                 & 58.70   \\ \hline% \cline{1-1}
		EchoFusion-small        & 90.43                     & 90.19                    & 78.95                 & 90.36                 & 79.08                  & 47.73       \\ \hline \hline% \cline{1-1}
		Ours-depth64              & 98.99                     & 89.97                   & 88.38                 & 90.60                 & 88.51                  & \textbf{70.58}       \\ \hline% \cline{1-1}
		Ours-depth128             & \textbf{99.75}       & 90.65                   & 88.97                 & \textbf{99.63}   & \textbf{89.06}   & 61.44       \\
		 \hline
	\end{tabular}
	\label{ap}
\end{table}

We use the KITTI protocol to evaluate the Average Precision (AP) of all the models mentioned above, 
with Intersection over Union (IoU) thresholds set at 0.3, 0.5, and 0.7. 
We highlight the best result for each metric in bold. 
As shown in Table \ref{ap}, 
Ours-depth128 and Ours-depth64 significantly outperform EchoFusion-small in all AP metrics 
except for the BEV AP at a threshold of 0.5. 
Moreover, while our methods are slightly inferior to EchoFusion in the BEV AP at thresholds of 0.5 and 0.7, 
both Ours-depth128 and Ours-depth64 surpass EchoFusion in all the 3D AP metrics, 
except when the 3D AP threshold is 0.3, 
where Ours-depth64 is slightly inferior to EchoFusion. 
When the threshold is set at 0.3, 
both the BEV AP and 3D AP of Ours-depth128 exceed EchoFusion by more than 10\%.

We present the errors for the outputs of all models in Table \ref{err}. 
Using an evaluation method similar to that of NuScenes, 
we first match the output bounding boxes with the ground truth. 
We then calculate the average Euclidean distance between the centers of 
the output and ground truth (GT) bounding boxes as the Average Translation Error (ATE), 
the average ratio of their diagonal lengths as the Average Scaling Error (ASE), 
and the average difference in their angles as the Average Orientation Error (AOE). 
The table indicates that, 
except for the slightly larger angle error in Ours-depth128, 
all the error metrics of our methods are lower than those of EchoFusion and EchoFusion-small.

\begin{table}[h]
	\centering
	\renewcommand{\arraystretch}{1.5}
	\setlength{\tabcolsep}{4pt}
	\caption{Prediction Errors using different methods. 
		%		IoU thresholds of 0.3, 0.5, and 0.7 are used. 
		%		The best result for each metric is highlighted in bold.
	}
	\begin{tabular}{|c||c|c|c|}
		\hline
		Method           & ATE (meter)   & ASE (\%)       & AOE (degree)  \\ \hline  \hline
		EchoFusion       & 0.25          & 6.30          & 7.74          \\ \hline
		EchoFusion-small & 0.29          & 8.08          & 7.92          \\ \hline
		Ours-depth64     & \textbf{0.23} & 6.27          & \textbf{7.34} \\ \hline
		Ours-depth128    & \textbf{0.23} & \textbf{6.23} & 7.99          \\ \hline
	\end{tabular}
	\label{err}
\end{table}

We also visualize the projected bounding boxes of each model's output alongside the ground truth in the images, as shown in Fig. \ref{vis}. 
This indicates that, in most cases, all models produce similar predictions, 
but there are some subtle differences in certain scenarios. 
For instance, in the second row, 
the bounding boxes predicted by our methods appear more accurate in terms of size and angle. 
In the third row, EchoFusion fails to detect a distant object, whereas this issue is not observed with the other methods.

The results above demonstrate that 
our models significantly improve detection performance by incorporating depth images to enhance image features. 
It is also worth noting that our models are smaller compared to EchoFusion:  
our models use approximately 19 GiB of memory during training on an RTX 3090, 
while EchoFusion requires about 24 GiB. 
During inference, our models utilize 3.8 GiB, 
whereas EchoFusion uses 4.1 GiB.

\subsection{Discussion and Limitation}
As mentioned in Subsection \ref{depth_generation}, inspired by 3DRIMR \cite{3drimr}, 
in the absence of a depth sensor, 
we can generate corresponding depth images from radar spectra using a GAN-based depth generator. 
If we disregard network complexity, 
we could directly integrate the depth generator into the existing model, as shown in Fig.  \ref{our_arch}. 
Alternatively, to avoid adding extra parameters, 
we could first train the depth generator on a subset of the data 
and then load the pretrained depth generator into the existing model to generate the depth images. 
This approach is feasible, as the depth images generated in 3DRIMR \cite{3drimr} are of high quality. 
However, since the K-Radar dataset lacks depth images, 
we can only project point clouds from a 64-line and a 128-line LiDAR onto the images to create ground truth depth images. 
These depth images are very sparse, 
and combined with the limited data available (approximately 4,500 frames for training in our experiments), 
the current depth generator we have trained cannot output high-quality depth images. 
We aim to address this issue in the future.

% ================================================================================================
\section{Conclusion and Future Work}

In this work, we propose a novel 3D object detection algorithm based on the fusion of cameras and 4D millimeter-wave radar for multimodal perception. 
This approach aims to enhance detection performance, cost-effectiveness, and robustness in all-weather environments. 
We are the first to utilize depth-aware images and 4D millimeter-wave radar spectra as inputs for 3D object detection, 
and we propose generating depth images from radar spectra when depth sensors are unavailable. 
Experimental results demonstrate that our model maintains strong detection performance while requiring less memory and computational resources.

However, while numerous studies have shown that using raw radar data can yield better detection performance compared to radar point clouds, 
raw radar data also requires significant storage capacity and contains considerable noise. 
In the future, we aim to explore radar data preprocessing techniques to 
reduce the size of the radar spectrum and remove noise before inputting it into the model.

%\begin{thebibliography}{00}
%\bibitem{b1} G. Eason, B. Noble, and I. N. Sneddon, ``On certain integrals of Lipschitz-Hankel type involving products of Bessel functions,'' Phil. Trans. Roy. Soc. London, vol. A247, pp. 529--551, April 1955.
%\bibitem{b2} J. Clerk Maxwell, A Treatise on Electricity and Magnetism, 3rd ed., vol. 2. Oxford: Clarendon, 1892, pp.68--73.
%\bibitem{b3} I. S. Jacobs and C. P. Bean, ``Fine particles, thin films and exchange anisotropy,'' in Magnetism, vol. III, G. T. Rado and H. Suhl, Eds. New York: Academic, 1963, pp. 271--350.
%\end{thebibliography}
%
%You may put all reference items in a separate file, say myRef.bib, in bibTex format.
%
\bibliographystyle{IEEETran}
\bibliography{references} %change to your file name (with suffix .bib)

\begin{thebibliography}{10}
\providecommand{\url}[1]{#1}
\csname url@samestyle\endcsname
\providecommand{\newblock}{\relax}
\providecommand{\bibinfo}[2]{#2}
\providecommand{\BIBentrySTDinterwordspacing}{\spaceskip=0pt\relax}
\providecommand{\BIBentryALTinterwordstretchfactor}{4}
\providecommand{\BIBentryALTinterwordspacing}{\spaceskip=\fontdimen2\font plus
\BIBentryALTinterwordstretchfactor\fontdimen3\font minus
  \fontdimen4\font\relax}
\providecommand{\BIBforeignlanguage}[2]{{%
\expandafter\ifx\csname l@#1\endcsname\relax
\typeout{** WARNING: IEEEtran.bst: No hyphenation pattern has been}%
\typeout{** loaded for the language `#1'. Using the pattern for}%
\typeout{** the default language instead.}%
\else
\language=\csname l@#1\endcsname
\fi
#2}}
\providecommand{\BIBdecl}{\relax}
\BIBdecl

\bibitem{4dsurvey}
L.~Fan, J.~Wang, Y.~Chang, Y.~Li, Y.~Wang, and D.~Cao, ``4d mmwave radar for
  autonomous driving perception: A comprehensive survey,'' \emph{IEEE
  Transactions on Intelligent Vehicles}, vol.~9, no.~4, pp. 4606--4620, 2024.

\bibitem{sls}
H.~Yang and J.~Wang, ``Sidelobe suppression of sar images by spectrum
  shaping,'' pp. 1668--1671, 2022.

\bibitem{cfar}
H.~Rohling, ``Radar cfar thresholding in clutter and multiple target
  situations,'' \emph{IEEE Transactions on Aerospace and Electronic Systems},
  vol. AES-19, no.~4, pp. 608--621, 1983.

\bibitem{dpft}
F.~Fent, A.~Palffy, and H.~Caesar, ``Dpft: Dual perspective fusion transformer
  for camera-radar-based object detection,'' \emph{arXiv preprint
  arXiv:2404.03015}, 2024.

\bibitem{exploringradardatarepresentations}
\BIBentryALTinterwordspacing
S.~Yao, R.~Guan, Z.~Peng, C.~Xu, Y.~Shi, W.~Ding, E.~G. Lim, Y.~Yue, H.~Seo,
  K.~L. Man, J.~Ma, X.~Zhu, and Y.~Yue, ``Exploring radar data representations
  in autonomous driving: A comprehensive review,'' 2024. [Online]. Available:
  \url{https://arxiv.org/abs/2312.04861}
\BIBentrySTDinterwordspacing

\bibitem{echofusion}
Y.~Liu, F.~Wang, N.~Wang, and Z.~Zhang, ``Echoes beyond points: Unleashing the
  power of raw radar data in multi-modality fusion,'' 2023.

\bibitem{polarformer}
Y.~Jiang, L.~Zhang, Z.~Miao, X.~Zhu, J.~Gao, W.~Hu, and Y.-G. Jiang,
  ``Polarformer: multi-camera 3d object detection with polar transformer,'' in
  \emph{AAAI}, ser. AAAI'23/IAAI'23/EAAI'23.\hskip 1em plus 0.5em minus
  0.4em\relax AAAI Press, 2023.

\bibitem{texas}
C.~Iovescu and S.~Rao, ``The fundamentals of millimeter wave radar sensors,''
  pp. 1--9, 2021.

\bibitem{nuscenes}
H.~Caesar, V.~Bankiti, A.~H. Lang, S.~Vora, V.~Liong, Q.~Xu, A.~Krishnan,
  Y.~Pan, G.~Baldan, and O.~Beijbom, ``nuscenes: A multimodal dataset for
  autonomous driving,'' pp. 11\,618--11\,628, 2020.

\bibitem{RADDet}
A.~Zhang, F.~E. Nowruzi, and R.~Laganiere, ``Raddet: Range-azimuth-doppler
  based radar object detection for dynamic road users,'' pp. 95--102, 2021.

\bibitem{zendar}
M.~Mostajabi, C.~M. Wang, D.~Ranjan, and G.~Hsyu, ``High resolution radar
  dataset for semi-supervised learning of dynamic objects,'' pp. 450--457,
  2020.

\bibitem{RADIATE}
M.~Sheeny, E.~De~Pellegrin, S.~Mukherjee, A.~Ahrabian, S.~Wang, and A.~Wallace,
  ``Radiate: A radar dataset for automotive perception in bad weather.''\hskip
  1em plus 0.5em minus 0.4em\relax IEEE Press, 2021, p. 1–7.

\bibitem{CARRADADC}
A.~Ouaknine, A.~Newson, J.~Rebut, F.~Tupin, and P.~Perez, ``Carrada dataset:
  Camera and automotive radar with range- angle- doppler annotations,'' Los
  Alamitos, CA, USA, pp. 5068--5075, 2021.

\bibitem{vod}
A.~Palffy, E.~Pool, S.~Baratam, J.~F.~P. Kooij, and D.~M. Gavrila,
  ``Multi-class road user detection with 3+1d radar in the view-of-delft
  dataset,'' \emph{IEEE Robotics and Automation Letters}, vol.~7, no.~2, pp.
  4961--4968, 2022.

\bibitem{astyx}
M.~Meyer and G.~Kuschk, ``Automotive radar dataset for deep learning based 3d
  object detection,'' pp. 129--132, 2019.

\bibitem{radar_scenes}
O.~Schumann, M.~Hahn, N.~Scheiner, F.~Weishaupt, J.~F. Tilly, J.~Dickmann, and
  C.~Wöhler, ``Radarscenes: A real-world radar point cloud data set for
  automotive applications,'' pp. 1--8, 2021.

\bibitem{k-radar}
D.-H. Paek, S.-H. Kong, and K.~T. Wijaya, ``K-radar: 4d radar object detection
  for autonomous driving in various weather conditions,'' \emph{Advances in
  Neural Information Processing Systems}, vol.~35, pp. 3819--3829, 2022.

\bibitem{centerfusion}
R.~Nabati and H.~Qi, ``Centerfusion: Center-based radar and camera fusion for
  3d object detection,'' pp. 1526--1535, 2021.

\bibitem{rcbevdet}
Z.~Lin, Z.~Liu, Z.~Xia, X.~Wang, Y.~Wang, S.~Qi, Y.~Dong, N.~Dong, L.~Zhang,
  and C.~Zhu, ``Rcbevdet: Radar-camera fusion in bird's eye view for 3d object
  detection,'' pp. 14\,928--14\,937, June 2024.

\bibitem{bevfusion}
Z.~Liu, H.~Tang, A.~Amini, X.~Yang, H.~Mao, D.~L. Rus, and S.~Han, ``Bevfusion:
  Multi-task multi-sensor fusion with unified bird's-eye view representation,''
  pp. 2774--2781, 2023.

\bibitem{lxl}
W.~Xiong, J.~Liu, T.~Huang, Q.-L. Han, Y.~Xia, and B.~Zhu, ``Lxl: Lidar
  excluded lean 3d object detection with 4d imaging radar and camera fusion,''
  \emph{IEEE Transactions on Intelligent Vehicles}, vol.~9, no.~1, p. 79–92,
  2024.

\bibitem{rcfusion}
L.~Zheng, S.~Li, B.~Tan, L.~Yang, S.~Chen, L.~Huang, J.~Bai, X.~Zhu, and Z.~Ma,
  ``Rcfusion: Fusing 4-d radar and camera with bird’s-eye view features for
  3-d object detection,'' \emph{IEEE Transactions on Instrumentation and
  Measurement}, vol.~72, pp. 1--14, 2023.

\bibitem{resnet}
K.~He, X.~Zhang, S.~Ren, and J.~Sun, ``Deep residual learning for image
  recognition,'' pp. 770--778, 2016.

\bibitem{fpn}
T.-Y. Lin, P.~Dollár, R.~Girshick, K.~He, B.~Hariharan, and S.~Belongie,
  ``Feature pyramid networks for object detection,'' pp. 936--944, 2017.

\bibitem{3drimr}
Y.~Sun, Z.~Huang, H.~Zhang, Z.~Cao, and D.~Xu, ``3drimr: 3d reconstruction and
  imaging via mmwave radar based on deep learning,'' pp. 1--8, 2021.

\bibitem{rgbd_survey}
A.~Lopes, R.~Souza, and H.~Pedrini, ``A survey on rgb-d datasets,''
  \emph{Computer Vision and Image Understanding}, vol. 222, p. 103489, 2022.

\bibitem{r2p}
Y.~Sun, H.~Zhang, Z.~Huang, and B.~Liu, ``R2p: A deep learning model from
  mmwave radar to point cloud,'' p. 329–341, 2022.

\bibitem{multi-obj}
Y.~Sun, Z.~Huang, H.~Zhang, and X.~Liang, ``3d reconstruction of multiple
  objects by mmwave radar on uav,'' pp. 491--495, 2022.

\end{thebibliography}

\end{document}